\pdfoutput=1

\documentclass[11pt]{article}

\usepackage{EMNLP2022}

\usepackage{times}
\usepackage{xcolor}

\usepackage[most]{tcolorbox}
\usepackage{latexsym}
\usepackage{graphicx}
\usepackage{booktabs}
\usepackage{multirow}
\usepackage{verbatim}

\usepackage{bbold}
\usepackage{xcolor}

\usepackage[ruled,vlined,linesnumbered]{algorithm2e}    
\usepackage{algpseudocode} 
\usepackage{multirow}

\usepackage{siunitx}
\usepackage{colortbl}
\usepackage{makecell}

\usepackage{subcaption}
\usepackage{bbm}
\usepackage{amssymb}

\definecolor{maroon}{cmyk}{0,0.87,0.68,0.32}

\usepackage[T1]{fontenc}

\usepackage[utf8]{inputenc}

\usepackage{microtype}

\usepackage{inconsolata}

\catcode`\@ = 12
\title{Precisely the Point: Adversarial Augmentations for  Faithful and Informative Text Generation}

\author{ \parbox{\linewidth}{\centering Wenhao Wu\textsuperscript{1}\thanks{\ \ Work is done during an internship at Baidu Inc.}, Wei Li\textsuperscript{2}, Jiachen Liu\textsuperscript{2}, Xinyan Xiao\textsuperscript{2},  Sujian Li\textsuperscript{1}\thanks{\ \ Corresponding author.}, Yajuan Lyu\textsuperscript{2} }\\
    \textsuperscript{1}Key Laboratory of Computational Linguistics, MOE, Peking University \\
  \textsuperscript{2}Baidu Inc., Beijing, China \\
  \texttt{\{waynewu,lisujian\}@pku.edu.cn}\\
  \texttt{\{liwei85,liujiachen,xiaoxinyan, lvyajuan\}@baidu.com}\\}

\begin{document}
\maketitle
\begin{abstract}
Though model robustness has been extensively studied in language understanding, the robustness of Seq2Seq generation remains understudied.
In this paper, we conduct the first quantitative analysis on the robustness of pre-trained Seq2Seq models. 
We find that even current SOTA pre-trained Seq2Seq model (BART) is still vulnerable, which leads to significant degeneration in faithfulness and informativeness for text generation tasks.
This motivated us to further propose a novel adversarial augmentation framework, namely AdvSeq, for generally improving faithfulness and informativeness of Seq2Seq models via enhancing their robustness. 
AdvSeq automatically constructs two types of adversarial augmentations during training, including implicit adversarial samples by perturbing word representations and explicit adversarial samples by word swapping, both of which effectively improve Seq2Seq robustness.
Extensive experiments on three popular text generation tasks demonstrate that AdvSeq significantly improves both the faithfulness and informativeness of Seq2Seq generation under both automatic and human evaluation settings.
\end{abstract}
\section{Introduction}

Recently, text generation has made significant progress thanks to the development of pre-trained sequence-to-sequence (Seq2Seq) models \cite{DBLP:conf/acl/LewisLGGMLSZ20, DBLP:conf/icml/ZhangZSL20,DBLP:conf/eacl/RollerDGJWLXOSB21}.
Despite being able to generate fluent and grammatical text, current state-of-the-art models tend to generate low-informative and unfaithful outputs that hallucinated with the given inputs \cite{welleck-etal-2019-dialogue,DBLP:conf/acl/MaynezNBM20}. 
One major reason is  that Seq2Seq models, generally trained with Negative Log-Likelihood (NLL), are still not robust enough to handle perturbations in the input \cite{DBLP:conf/acl/KangH20}.
This vulnerability can lead to significant degeneration of the generalization performance for text generation tasks \cite{DBLP:conf/acl/ChengJM19}. 
Table \ref{tab:adv_sample} provides a typical example to show the vulnerability of the model to perturbations.
We can see that the words ``aeroplane'' and ``injuries'' in the original text are revised as ``aeroplanes'' and ``accident'' respectively, which do not alter the  meaning of the input. However, the output generated by BART obviously hallucinates the input.
\begin{table}[t]
    \centering
     \begin{tabular}{p{0.46\textwidth}}
      \Xhline{2\arrayrulewidth}
  \textbf{Original Input $x$}: \, Police said the incident happened  near the village of Tockwith. North Yorkshire Police said it was believed the \textcolor{blue}{aeroplane} had suffered engine failure. Sgt Andy Graham, who attended the crash, praised the "great piloting skills resulting in no \textcolor{blue}{injuries}".\\
  \textbf{Output from $x$}: \, A light aircraft has crashed into a field in Yorkshire, but no-one was injured.\\
  \Xhline{2\arrayrulewidth}
  \textbf{Adversarial Input $x'$:} \,  Police said the incident happened near the village of Tockwith. North Yorkshire Police said it was believed the \textcolor{blue}{aeroplanes} had suffered engine failure. Sgt Andy Graham, who attended the crash, praised the "great piloting skills resulting in no \textcolor{blue}{accident}".\\
  \textbf{Output from $x'$}:\, Two aeroplanes have crashed in North Yorkshire, \textcolor{red}{with one of them landing safely on its roof}.\\
     \Xhline{2\arrayrulewidth}
\end{tabular}
    \caption{A sample of the XSum dataset for text summarization generated by fine-tuned BART, where minor perturbations (words in blue) cause low-informative and unfaithful generation (span in red).}
    \label{tab:adv_sample}
\end{table}

The model robustness problem has been extensively studied in language understanding, however, it is still understudied in language generation, especially for pre-trained Seq2Seq models \cite{li2021pretrained}.
The popular adversarial learning methods for language understanding, such as  FreeLB \cite{DBLP:conf/iclr/ZhuCGSGL20}, are not effective on Seq2Seq generation tasks (as shown in Table~\ref{tab:result_sum}).
Although few previous work have attempted to craft adversarial samples for Seq2Seqs on machine translation~\cite{DBLP:conf/iclr/BelinkovB18, DBLP:conf/aaai/ChengYCZH20}, they have not been extensively studied across various generation tasks.
Furthermore, they have not been studied on current pre-trained Seq2Seq models.

To address this problem, we provide the first quantitative analysis on the robustness of pre-trained Seq2Seqs.
Specifically, we quantitatively analyze the robustness of BART across three generation tasks, i.e.,  text summarization, table-to-text, and dialogue generation.
Through slightly modifying the input content, we find that the corresponding output significantly drops in both informativeness and faithfulness, which also demonstrate the close connection between the robustness of Seq2Seq models and their informativeness and faithfulness on generation tasks.

Based on the analysis above, we further propose a novel \textbf{Adv}ersarial augmentation framework for  \textbf{Seq}uence-to-Sequence generation (AdvSeq) to enhance its robustness against perturbations and thus obtain an informative and faithful text generation model.
AdvSeq constructs challenging and factually consistent adversarial samples and learns to defend against their attacks.
To increase the diversity of the adversarial samples,  AdvSeq applies two types of perturbation strategies, implicit adversarial samples (AdvGrad) and explicit token swapping (AdvSwap), efficiently utilizing the  back-propagate gradient during training.
AdvGrad directly perturbs word representations  with gradient vectors, while AdvSwap utilizes gradient directions for searching for token replacements.
To alleviate the vulnerability  of NLL, AdvSeq adopts a KL-divergence-based loss function to train with those adversarial augmentation samples, which promotes higher invariance in the word representation space \cite{DBLP:journals/pami/MiyatoMKI19}.


We evaluate AdvSeq by extensive experiments on three generation tasks: text summarization, table-to-text, and dialogue generation.
Our experiments demonstrate that AdvSeq can effectively improve Seq2Seq robustness against adversarial samples, which result in better informativeness and faithfulness on various text generation tasks.
Comparing to existing adversarial training methods for language understanding and data augmentation methods for Seq2Seqs, AdvSeq can more effectively improve both the informativeness and faithfulness for text generation tasks.

We summarize our contributions as follows:
\begin{itemize}
\item To the best of our knowledge, we are the first to conduct quantitative analysis on the robustness of pre-trained Seq2Seq models, which reveal its close connection with their informativeness and faithfulness on generation tasks.
\item We propose a novel adversarial argumentation framework for Seq2Seq models, namely AdvSeq, which effectively improves their informativeness and faithfulness on various generation tasks via enhancing their robustness.
\item Automatic and human evaluations on three popular text generation tasks validate that AdvSeq significantly outperforms several strong baselines in both informativeness and faithfulness.
\end{itemize}

\section{ Seq2Seq Robustness Analysis}\label{sec:robust_ana}

In this section, we analyze the robustness of the Seq2Seq  by evaluating its performance on adversarial samples.
In brief, after the input contexts are minorly modified, we check whether the model maintains its informativeness and faithfulness. 
A robust model should adaptively generate high-quality texts corresponding to the modified inputs.

Following the definition of adversarial examples on Seq2Seq models, \textit{adversarial examples should be meaning-preserving on the source side, but meaning-destroying on the target side} \cite{DBLP:conf/naacl/MichelLNP19}.
Formally,  given an input context $x$ and its reference text  $y_{ref}$  from the test set of a task, and a Seq2Seq model $f_\theta$ trained on the training set, we fist collect the original generated text   $y = f_\theta(x)$.
We measure its faithfulness and informativeness by $E_f(x,y)$ and $E_i(x,y,y_{ref})$, where $E_{f}$ and $E_{i}$ are the faithfulness and informativeness  metrics, respectively.
Then, we craft an adversarial sample $x'$ by slightly modifying $x$ trying not to alter its original meaning and generate $y'$ grounded on $x'$.
Finally, we measure the \textit{
target relative score decrease} \cite{DBLP:conf/naacl/MichelLNP19} of faithfulness  after attacks by:
\begin{equation}
    d = \frac{E_f(x,y)-E_f(x',y')}{E_f(x,y)}
\end{equation}
We calculate the decrease of informativeness similarly.
We also report the entailment score of $x'$ towards $x$: $S(x, x')$ to check whether the  modification  changes the meaning.

\paragraph{Evaluation Settings} We apply  BART as $f_\theta$ and conduct evaluations on three datasets, XSum for text summarization, WIKIPERSON for table-to-text, and Dialogue NLI for dialogue generation, with 2,000 samples sampled from each dataset.
Evaluation metrics for different tasks are listed in Table \ref{tab:adv_metric}, and the details  are introduced in \S \ref{sec:exp}.

As a preliminary study of robustness, we design a simple word swapping-based method for crafting adversarial samples.
For word $w_i \in x$, we first calculate  its salience score  as  $f_\theta(y_{ref}|x)-f_\theta(y_{ref}|x\setminus w_i)$, where $f_\theta(y|x)$ is the validation score of generating $y$ given $x$, and $x\setminus w_i$ is input $x$ with word $w_i$ deleted.
We then sort the salient score to get the swapping orders of words in $x$,  giving priority to those with higher scores.
Following the orders, we iteratively swap each word with its top 10 nearest neighbors in the word embedding space and keep the replacement if the output BLEU score decreases  after swapping.
For better meaning preservation, we hold the maximum difference in edit distance to a constant of 30 for each sample.
The details of algorithm and  evaluation metrics are introduced in Appendix \ref{app:robust_analyse}.

\paragraph{Attack Results} Reported  in Table \ref{tab:adv_result}, for informativeness, text summarization drops by 8.75\% in ROUGE-L, dialogue generation increases by 16.48\% on the  reference perplexity, table-to-text drops by 4.41\% and 6.11\% on the PATENT recall and F-1, respectively.
For faithfulness, text summarization drops by 15.53\% and 5.67\% in FactCC and QuestEval, table-to-text generation drops in PARENT precision by 6.24\%, dialogue generation increase in the perplexity of entailed candidates by $5.67\%$.
Overall, BART is still not robust enough to  tackle minor perturbations  in the input, which lead to degeneration of the
generalization performance of both informativeness and faithfulness.

\begin{table}[]
    \centering
    
\setlength\tabcolsep{2.8pt}
\renewcommand{\arraystretch}{0.9}
    \begin{tabular}{l|l l l l}
    \Xhline{2\arrayrulewidth}
         Task&$S$&$E_i$&$E_f$\\
         \Xhline{2\arrayrulewidth}
          Summarization & EntS &ROURGE-L &CC/QE \\
          Table-to-text&-&PARENT&PARENT\\
          Dialogue NLI&EntS&PPL&PPL/Ranking\\
    \Xhline{2\arrayrulewidth}
    \end{tabular}
    \caption{Evaluation metrics for different tasks, where we apply entailment score (EntS) for $S$, FactCC (CC), QuestEval (QE) for $E_f$ of summarization and ROUGE-L (R-L) for $E_i$ of summarization. PPL is the abbreviation for perplexity.}
    \label{tab:adv_metric}
\end{table}
\begin{table}[t]
    \begin{subtable}[h]{0.5\textwidth}
    \renewcommand{\arraystretch}{0.9}

    \centering    
    \setlength\tabcolsep{11pt}
    \begin{tabular}{c|ccccc}
    \Xhline{2\arrayrulewidth}
    \multirow{2}{*}{Input}&\multicolumn{3}{c}{PARENT}\\
    
    &Precision&Recall&F-1\\
    \Xhline{2\arrayrulewidth}
        Ori.&57.61&97.54&71.56\\
        Adv.&54.01&93.17&67.19\\
         \Xhline{2\arrayrulewidth}
        d&$6.24^{\dagger}$&$4.41^{\dagger}$&$6.11^{\dagger}$\\
    \Xhline{2\arrayrulewidth}
    \end{tabular}
    \caption{Table-to-text (WIKIPERSON)}
    \label{tab:my_label}
    
    \end{subtable}
             \renewcommand{\arraystretch}{0.9}

     \begin{subtable}[h]{0.5\textwidth}
     \centering    
    \setlength\tabcolsep{5pt}
    \begin{tabular}{c|ccccc}
    \Xhline{2\arrayrulewidth}
    
    \multirow{2}{*}{Input}&\multicolumn{3}{c}{Perplexity}&\multirow{2}{*}{Hit@1$\uparrow$}\\
    &Ref$\downarrow$&Ent.$\downarrow$&Con.$\uparrow$&\\
    \Xhline{2\arrayrulewidth}
        Ori.&10.30&22.08&16.15&37.64 \\
        Adv.&12.01&22.75&16.60&35.62\\
      \Xhline{2\arrayrulewidth}
        d&$-16.48^{\dagger}$&$-2.95^{\dagger}$&0.25&$5.67^{\dagger}$\\
    \Xhline{2\arrayrulewidth}
    \end{tabular}
    \caption{Dialogue Generation (Dialogue NLI)}
    \label{tab:my_label}
    \end{subtable}
    
             \renewcommand{\arraystretch}{0.9}

    \begin{subtable}[h]{0.5\textwidth}
    \centering    

      \setlength\tabcolsep{8.5pt}
    \begin{tabular}{c|cccccc}
    \Xhline{2\arrayrulewidth}
     &EntS&R-L&CC.&QE\\

    \Xhline{2\arrayrulewidth}
   
        Ori.& -&35.96&16.74& 43.46\\
        Adv.& 83.6&32.83&$14.18$&41.20\\
        \Xhline{2\arrayrulewidth}
        d&-&$8.75^{\dagger}$&$15.53^{\dagger}$&$5.20^{\dagger}$\\
    \Xhline{2\arrayrulewidth}
    \end{tabular}
    \caption{Text Summarization (XSum)}
    
    \end{subtable}
    \caption{Evaluation performance of fine-tuned BART on 
    original samples (Ori.) vs adversarial samples (Adv.), $d$ is the target
relative score decrease  of every metric. $\dagger$: significantly decrease (p < 0.01) by t-test.}
    \label{tab:adv_result}
\end{table}

\section{AdvSeq Framework}
Inspired by the findings in the previous section, we propose to improve the informativeness and faithfulness of a Seq2Seq model via enhancing its robustness.
We propose our framework AdvSeq for robustly fine-tuning pre-trained Seq2Seq models.
During training, AdvSeq utilizes  gradient information to automatically construct challenging but factually consistent augmentation samples.
For better diversity,  we construct two kinds of augmentation samples,  implicit adversarial sample (AdvGrad) and explicit token replacement (AdvSwap).
In the following, we introduce AdvSeq in detail.

Given an input sample $(x,y)$, a Seq2Seq model with parameters $\theta$ learns to generate fluent text by training with NLL loss.
Considering the vulnerability of NLL, we further measure and optimize probability distribution changes caused by small random perturbations via KL divergence.
The overall loss function w.r.t. the clean sample $(x,y)$ is then defined as:
\begin{align}
    &\mathcal{L}_{o}(x+\delta_x,y+\delta_y, \theta) =-\sum_{y_t\in y} \log p(y_t|x,y_{<t}) \   \\
    & + KL_S  (p(y_t|y_{<t}+\delta_y,x+ \delta_x),p(y_t|y_{<t},x)) \nonumber
    \label{eq:KL_loss}
\end{align}
where the first term is NLL, $KL_S$ \footnote{$KL_S(x,y) = KL(x|y)+KL(y|x)$} is the symmetry of KL divergence,  and perturbations $\delta_x, \delta_y$ are sampled from a uniform distribution and added on word embeddings as default.
\textit{After that,  we back-propagate $\mathcal{L}_{o}$ and apply its gradient  to construct AdvGrad and AdvSwap.}

\paragraph{AdvGrad} Directly utilizing back-propagated gradient of  $\mathcal{L}_o$ as perturbations, we construct implicit adversarial samples.     
Instead of randomly perturbing word representations  like $\mathcal{L}_{o}$,  AdvGrad further searches for stronger perturbations that mostly affect the generation process   by solving:
\begin{align}
&\min_\theta \mathbbm{E}_{(x,y)}[\max_{\delta_x,\delta_y}\mathcal{L}_o(x+\delta_x,y+\delta_y,\theta)] \\ \label{eq:adv_solve}
&s.t. ||\delta_x||_F\leq \epsilon, \, ||\delta_y||_F\leq \epsilon \nonumber
\end{align}
where we constrain the Frobenius norm of a sequence by $\epsilon$.
Because it is intractable to solve this equation, we perform   gradient ascent \cite{DBLP:conf/iclr/MadryMSTV18} on the  $\delta_x$ to  approximate the solution:
\begin{align}
    \delta^{'}_x &= \Pi_{||\delta_x||_F  \leq  \epsilon} (\delta_x + \alpha*g_x/||g_x||_F) 
    \\
     g_x&=\nabla_{\delta_x}\mathcal{L}_o(x+\delta_x,y+\delta_y)
     \label{eq:adv_update}
\end{align}
where $\delta{'}_x$ is the updated adversarial perturbation, $\alpha$ is the learning rate of the update, $\Pi_{||\delta_x||_F \leq \epsilon}$ performs a projection onto the $\epsilon$-ball.
$\delta^{'}_y$ is approximated though similar steps.
Though previous works apply multi-step updates  for a closer approximation \cite{DBLP:conf/iclr/MadryMSTV18,DBLP:conf/iclr/ZhuCGSGL20}, we find one-step update is effective  and efficient enough for AdvGrad.
By perturbing the word embeddings with the adversarial perturbations: $x'= x+\delta^{'}_x$, $y'= y+\delta^{'}_y$, we get a pair of parallel augmentation samples, $(x^\prime,y^\prime)$.
Because the perturbations are implicit and  minor (within a $\epsilon$-ball ), we consider $(x^\prime,y^\prime)$ to be meaning preserving.
We then train with $(x^\prime,y^\prime)$  by optimizing  the KL divergence between its output word distributions $p(y'|x')$  with the original $p(y|x)$ by:
\begin{equation}
    \mathcal{L}_i = \sum_{y_t\in y}KL_S (p(y_t|y^\prime_{<t},x^\prime),p(y_t|y_{<t},x)) \label{eq:adv_grad}
\end{equation}

\paragraph{AdvSwap} Simultaneously utilizing gradient directions of $\mathcal{L}_o$, we  construct explicit adversarial samples by token swapping.
The core idea  is to identify and swap salient tokens in $x$ without or minorly changing the original meaning of $x$. 
The first procedure is to identify a set of salient tokens in $x$ that attributed most to  the generation.
We formulate this procedure as a causal inference problem \cite{DBLP:journals/tkdd/YaoCLLGZ21, DBLP:conf/emnlp/Xie00LD21}.
Concretely, given target context $\mathcal{Y}$, which can either be the whole reference text $y$ or sampled spans from it,  \textit{we need to infer a set of most relevant tokens $\mathcal{X}$  from $x$ that contribute most for generating $\mathcal{Y}$}.
Because the difficulties of this procedure  depend on specific tasks, we apply two strategies for searching  $\mathcal{X}$ in different tasks:
\begin{itemize}
\item \textbf{Gradient Ranking}: Select tokens $\{x_t$ $\in x\}$ with highest $k\%$  two-norm of gradient  $||\nabla_{x_i}\mathcal{L}_o(x,\mathcal{Y})||$.
\item \textbf{Word Overlapping}: Find  tokens in $x$ that overlap with $\mathcal{Y}$:  $ x\cap\mathcal{Y}$.
\end{itemize}
For tasks like table-to-text that word overlapping is enough to infer the precise causality, we randomly sample several spans from $y$ as $\mathcal{Y}$ and find the corresponding $\mathcal{X}$ by word overlapping.
While, for highly abstractive generation like text summarization, we use the global information  $y$ as $\mathcal{Y}$ and infer $\mathcal{X}$ by gradient ranking, which measures the salience of a token by gradient norm \cite{DBLP:journals/corr/SimonyanVZ13,DBLP:conf/naacl/LiCHJ16}.

After the $\mathcal{X}$ is inferred, we search around the neighbors of word embedding space for meaning preserving swapping, utilizing the semantic textual similarity of word embeddings  \cite{li-etal-2020-sentence}.
To make the adversarial samples more challenging, we search at the direction of  gradient ascent to replace $x_t \in \mathcal{X}$ with $\hat{x_t}$:
\begin{equation}
\hat{x}_t = \operatorname*{argmax}_{x_i \neq x_t} \cos(e_{x_i}-e_{x_t}, \nabla_{x_i}\mathcal{L}_o(x,\mathcal{Y})) \label{eq:swap}
\end{equation}
where  $e_{x_i}$ is the word embedding of $x_i$ in the vocabulary list, and $\cos(\cdot,\cdot)$ is the cosine similarity of two vectors.
After word swapping  is done for all the tokens in $\mathcal{X}$, we get the adversarial sample $x''$.
We train the explicit adversarial sample $(x'',\mathcal{Y}) $ with KL divergence:
\begin{equation}
    \mathcal{L}_e = \sum_{y_t\in\mathcal{Y}}KL_S (p(y_t|y_{<t},x''),p(y_t|y_{<t},x))
    \label{eq:adv_swap}
\end{equation}

\paragraph{Overall Training} For efficiently utilizing  the first  back-propagate step of $\mathcal{L}_o$, we apply the “Free” Large-Batch Adversarial Training (FreeLB) \cite{DBLP:conf/iclr/ZhuCGSGL20}.
For every training step, we first forward and back-propagate  $\mathcal{L}_o$ for  constructing AdvGrad and AdvSwap  while saving the gradient $\nabla_\theta \mathcal{L}_o$ with respect to  $\theta$.
Next, we forward and  back-propagate the loss function  of two  augmentations:  $\mathcal{L}_i + \mathcal{L}_e$ and  accumulate its gradient with respect to model  parameters $\theta$.
Finally, we update  $\theta$ with previous two-step accumulated gradient.
The overall training procedure is summarized in the Algorithm~\ref{app:advseq_al} in the appendix.


\begin{table*}[t]
\centering
    \setlength{\tabcolsep}{2.5mm}
  \begin{tabular}{l|ccccc|ccccc}
   \Xhline{2\arrayrulewidth}
    
      \multirow{2}{*}{Datasets}&\multicolumn{5}{c}{\textbf{XSum}}  &
      \multicolumn{5}{|c}{\textbf{CNN/DM}} \\
      & R-1 & R-2 & R-L &CC&QE& R-1 & R-2 & R-L &CC&QE\\
      
   \Xhline{2\arrayrulewidth}
   BART&45.20&21.90&36.88&23.64&45.89&44.08&20.92&40.79&76.09&51.15\\
   FreeAT& 45.51&22.08&37.16&23.23&45.16&44.02&20.88&40.75&\textbf{77.69}&51.09\\
   FreeLB&45.25&21.98&36.89&23.13&45.90&44.16&21.05&40.83&76.53&51.18\\
   \Xhline{2\arrayrulewidth}
   LT&44.42&21.10&36.17&23.99&45.29&44.05&20.89&40.79&\underline{77.06}&50.97\\
   CLIFF&44.63&21.39&36.43&23.51&45.45&44.29&21.14&41.02&75.66&50.47\\
  
   R3F&45.47&22.02&37.24&\textbf{26.10}&45.83&44.31&21.14&41.00&76.64&51.14\\
   SSTIA&45.82&22.42&37.50&23.89&46.05&\underline{44.76}&21.46&\underline{41.57}&71.69&51.20\\
   \Xhline{2\arrayrulewidth}
   \Xhline{2\arrayrulewidth}
   AdvGrad&\textbf{46.13}&\textbf{22.72}&\textbf{37.81}&23.49&$\textbf{47.82}^{\dagger}$&44.57&\underline{21.50}&41.36&75.06&50.67\\
   AdvSwap&45.49&22.21&37.21&\underline{24.70}&$47.25^{\dagger}$&44.63&21.47&41.35&74.46&\underline{51.21}\\
   AdvSeq&\underline{46.06}&\underline{22.65}&\underline{37.69}&24.67&$\underline{47.61^{\dagger}}$&\textbf{44.96}&\textbf{21.73}&\textbf{41.79}&72.43&$\textbf{51.50}^{\dagger}$\\
    \Xhline{2\arrayrulewidth}
   \end{tabular}
  \caption{Experimental results on text summarization, where R-1 (ROUGE-1), R-2 (ROUGE-2), R-L (ROUGE-L) are informative metrics and CC, QE report faithfulness. The last block reports the results of our methods. The underlines indicate the second-best performance. $\dagger$: significantly better than all the baseline model (p < 0.01).}
  \label{tab:result_sum}
\end{table*}

\begin{table}[t]
    \centering
        \setlength{\tabcolsep}{2.5mm}
      \renewcommand{\arraystretch}{0.9}

    \begin{tabular}{l|cc}

    \Xhline{2\arrayrulewidth}
         &BLEU&\textbf{PARENT}  \\
     \Xhline{2\arrayrulewidth}
    \citet{wang-etal-2020-towards}&24.56&53.06\\
    BART & \textbf{31.30}&56.40 \\
    LT&29.70&56.69\\
    Aug-plan&17.12&56.75\\
    R3F&31.08&56.56\\
    SSTIA&30.34&56.84\\
     \Xhline{2\arrayrulewidth}
     AdvGrad&\underline{31.17}&56.81\\
     AdvSwap&29.95&\underline{56.94}\\
    AdvSeq&30.37&$\textbf{57.33}^{\dagger}$\\
 \Xhline{2\arrayrulewidth}
    \end{tabular}
    \caption{Experimental results on WIKIPERSON. Note that the reference text in this dataset may contain hallucinated content, so BLEU scores can not measure the faithfulness of generation.}
    \label{tab:table-to-text}
\end{table}

\begin{table}[t]
    \centering
        \setlength{\tabcolsep}{2.5mm}
      \renewcommand{\arraystretch}{0.9}

    \begin{tabular}{l|cccc}
    \Xhline{2\arrayrulewidth}
    &\multicolumn{3}{c}{Perplexity}&\multicolumn{1}{c}{Hit@1}\\
    
    &Ref.$\downarrow$&Ent.$\downarrow$&Con.$\uparrow$&Ent.\%$\uparrow$\\
      \Xhline{2\arrayrulewidth}
    BART&10.9&\underline{22.2}&16.1&36.5\\
    LT&11.2&23.6&\underline{16.7}&38.0\\
    SSTIA&\underline{10.7}&24.3&\textbf{18.4}&\textbf{40.9}\\
    AdvSeq&$\textbf{9.8}^\dagger$&$\textbf{21.0}^\dagger$&15.9&\underline{38.7}\\ 

 \Xhline{2\arrayrulewidth}
    \end{tabular}
    \caption{Experimental results on Dialogue NLI, where Ref, Ent and Con indicates perplexity on reference, entailed  and contradict candidate utterances, respectively.}
    \label{tab:dialogue_nli}
\end{table}

\section{Experiment Setup}\label{sec:exp}

\subsection{Datasets}

We conduct experiments on four datasets of three text generation tasks: text summarization, table-to-text generation, and dialogue generation.
For text summarization,  we use XSum \cite{DBLP:conf/conll/NallapatiZSGX16} and CNN/DM \cite{hermann2015teaching} for evaluation.
For table-to-text generation, we use WIKIPERSON \cite{wang-etal-2018-describing}.
For dialogue generation, we apply dialogue NLI \cite{welleck-etal-2019-dialogue}.
Details of all datasets are listed in Table~\ref{tab:dataset}.

\subsection{Automatic Metric}

\paragraph{Informative Metric} We apply ROUGE  $F_1$ \cite{lin2004rouge} to evaluate text summarization, BLEU \cite{papineni-etal-2002-bleu} for table-to-text. 
For dialogue generation, given an example $(x, y)$ consists of a dialogue history $x = \{p_1, \dots , p_k, u_1, \dots , u_t\}$ and reference utterance $y$, where $p_i$ is a given persona sentence and $u_i$  a dialogue utterance, we report the perplexity of generating reference response $y$ to evaluate informativeness.

\paragraph{Faithfulness Metric} For text summarization, since previous works  \cite{pagnoni-etal-2021-understanding} report the unreliability of existing factual metrics, we report two different types of metrics, FactCC (CC) \cite{DBLP:conf/emnlp/KryscinskiMXS20} based on textual entailment and QuestEval (QE) \cite{scialom-etal-2021-questeval} based on question answering, for reliable comparison.
For table-to-text generation, we report the PARENT \cite{DBLP:conf/acl/DhingraFPCDC19} score of generated texts, which is a hybrid  measurement of its faithfulness and informativeness.
For dialogue generation, we report the perplexity of generating entailed and contradicted candidate utterances provided in the dialogue NLI.
We also report \textbf{Hit@1} of ranking candidate utterances by their modeling perplexity (the lower the better), which is the probability that the top one candidate is entailed by the dialogue history. 
Each example in dialogue NLI contains 10 entailment and 10 conflict utterances.

\subsection{Baseline Methods}
In all of our experiments, we employ the pre-trained Seq2seq model BART \cite{DBLP:conf/acl/LewisLGGMLSZ20} as our base model, which is then fine-tuned with various augmentation methods for particular tasks

\paragraph{Adversarial Training} While adversarial training has achieved promising results in natural language understanding, its performance on text generation lacks extensive evaluations.
We evaluate two representative methods, \textbf{FreeAT} \cite{shafahi2019adversarial} and \textbf{FreeLB} \cite{DBLP:conf/iclr/ZhuCGSGL20} on Seq2Seq tasks.

\paragraph{Faithfulness Augmentation} \textbf{CLIFF}     \cite{DBLP:conf/emnlp/Cao021} is a contrastive learning-based method which learns to discriminate various heuristically constructed augmentation samples.
Loss Truncation (\textbf{LT}) \cite{DBLP:conf/acl/KangH20} is a faithfulness augmentation method which adaptively removes noisy examples during training.
 \textbf{Aug-plan} \cite{DBLP:conf/aaai/0001ZCS21} augments  table-to-text training by incorporating auxiliary entity information.

\paragraph{Common Augmentation} We also compare AdvSeq with other common data augmentation methods for text generation. 
\textbf{R3F} \cite{aghajanyan2020better} fine-tunes pre-trained language models with random noises on embedding representations.
\textbf{SSTIA} \cite{xie2021target} constructs  the target-side augmentation sample  by a mix-up strategy.
\section{Implementation Details}\label{sec:imp_details}
During fine-tuning, we set the learning rate to  3e-5, label smoothing to 0.1.
For XSum and CNN/DM, we follow the parameter settings of \citet{DBLP:conf/acl/LewisLGGMLSZ20}.
For WIKIPERSON and Dialogue NLI we apply the similar parameter settings with \citet{DBLP:conf/acl/LewisLGGMLSZ20} on CNN/DM, except we do not apply trigram block during inference.
For AdvGrad we set $\alpha$ to 4e-1, $\epsilon$  to 2e-1.
For AdvSwap we randomly select two spans that contain overlapping tokens with $x$ as $\mathcal{Y}$ for table-to-text, we set $k$ to 0.15 for gradient ranking for text summarization and dialogue NLI.
\section{Results}
\subsection{Automatic Evaluation}

\begin{table}[t]
        \centering

     \setlength\tabcolsep{3.0pt}
      \renewcommand{\arraystretch}{0.9}

    \begin{subtable}[h]{0.5\textwidth}
    \begin{tabular}{l|cccccc}
    \Xhline{2\arrayrulewidth}
    &\multicolumn{3}{c}{Faithful.}& \multicolumn{3}{c}{Inform.}\\
   Model &Win$\uparrow$&Tie&Lose$\downarrow$&Win$\uparrow$&Tie&Lose$\downarrow$\\
    \Xhline{2\arrayrulewidth}
   SSTIA&20.0&63.5&\textbf{16.5}&23.5&52.0&24.5\\
   CLIFF&16.0&59.0&25.0&25.5&52.0&\textbf{22.5}\\
   AdvSeq&\textbf{27.0}&53.5&19.5&\textbf{33.5}&43.5&23.0\\
    \Xhline{2\arrayrulewidth}
    \end{tabular}
    \caption{XSum}
    \label{tab:my_label}
    \end{subtable}
    \begin{subtable}[h]{0.5\textwidth}
    \setlength\tabcolsep{2.5pt}
      \renewcommand{\arraystretch}{0.9}
    \begin{tabular}{l|cccccc}
    
    \Xhline{2\arrayrulewidth}
    &\multicolumn{3}{c}{Faithful.}& \multicolumn{3}{c}{Inform.}\\
   Model &Win$\uparrow$&Tie&Lose$\downarrow$&Win$\uparrow$&Tie&Lose$\downarrow$\\
   \Xhline{2\arrayrulewidth}
    SSTIA&25.0&57.5&\textbf{17.5}&21.0&64.0&\textbf{15.0}\\
   Aug-plan&10.5&22.0&67.5&18.0&27.0&55.0\\
   AdvSeq&\textbf{40.0}&41.0&19.0&\textbf{31.5}&52.5&16.0\\
    \Xhline{2\arrayrulewidth}
    \end{tabular}
    \caption{WIKIPERSON}
    \label{tab:my_label}
    \end{subtable}

    \caption{Percentage of generated text that are better than, tied with or worse than BART, in  faithfulness (Faithful.) and informativeness (Inform.). The Cohen’s kappa scores are 63.46 and 60.35 for two aspects on XSum, and 52.89 and 49.47 for WIKIPERSON.}
    \label{tab:human_eval}
\end{table}

\paragraph{Text Summarization}
Table \ref{tab:result_sum} reports results on XSum and CNN/DM.
As demonstrated by the results of FreeLB and FreeAT, directly fine-tuning with adversarial samples by NLL loss does not benefit BART.
Though they construct adversarial samples similarly based on gradient, AdvGrad achieves much better performance. 
Compared with all baselines in respective of informativeness, our models produce the best ROUGE scores.
Compared with all baselines in respective of faithfulness, our models also produce significantly better QE scores, which correlate better with human judgments than CC scores on both datasets, as claimed by \citet{DBLP:conf/emnlp/Cao021}. 
Specifically, both AdvGrad and AdvSwap improve ROUGE and QE scores across datasets.
Through combing them, AdvSeq produces a better performance on CNN/DM.
\paragraph{Table-to-text} 
Due to the existence of noisy training samples, PARENT  is considered a better metric than BLEU by matching the content with both the input table and reference text \cite{DBLP:conf/acl/DhingraFPCDC19}.
As reported in Table \ref{tab:table-to-text}, AdvSeq obtains significant higher PARENT scores comparing with all baselines.
Both AdvGrad and AdvSwap improve PARENT scores, and through combining them, AdvSeq further achieves the best performance.
\paragraph{Dialogue Generation} As reported in Table \ref{tab:dialogue_nli}, AdvSeq produces the lowest perplexity on  reference and entailed candidates, exceeding the second ranked systems by 8.4\% and 5.4\%, respectively. 
As AdvSeq mainly improves faithfulness by learning  from  factually consistent samples, it does not  help with distinguishing contradictory samples.
Thus, AdvSeq improves Hit@1 over BART and surpasses LT  by producing lower perplexity on entailed candidates without changing the perplexity of contracted candidates.
Overall, the results indicates that AdvSeq improves both informativeness and faithfulness of dialogue generation.

\begin{figure*}
    \centering
     \begin{subfigure}[b]{0.48\textwidth}
         \centering
         \includegraphics[width=\textwidth]{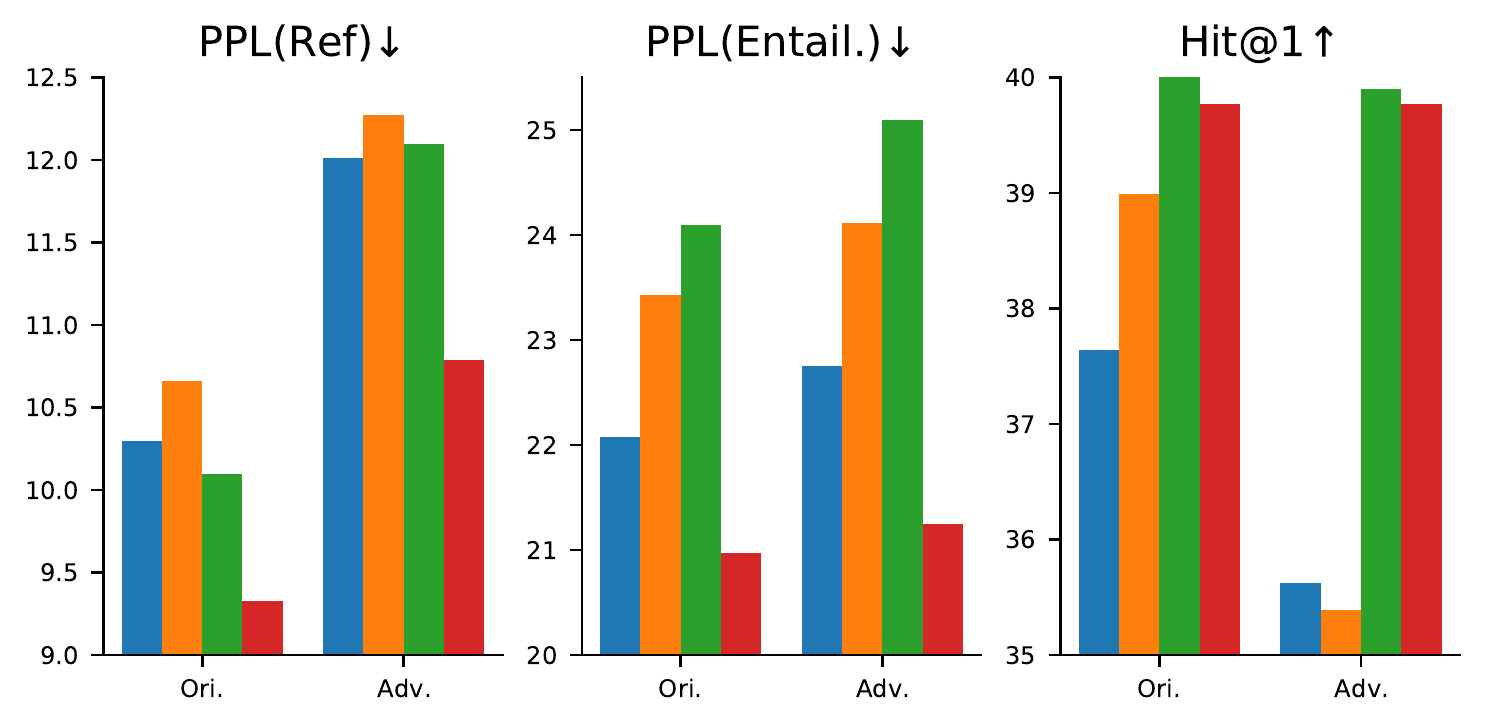}
         \caption{Dialogue generation (Dialogue NLI)}
          
     \end{subfigure}
     \begin{subfigure}[b]{0.48\textwidth}
         \centering
         \includegraphics[width=\textwidth]{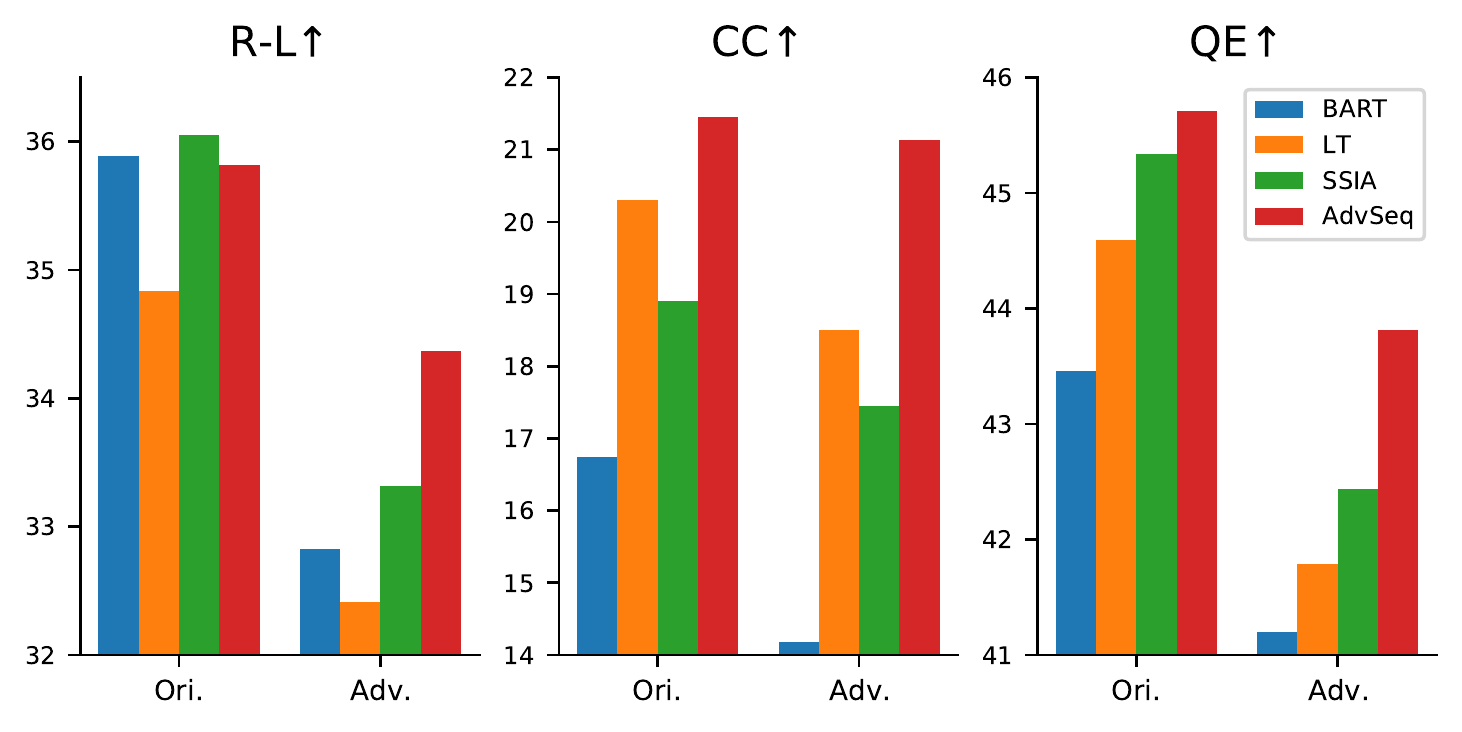}
         \caption{Text Summarization (XSum)}
        
     \end{subfigure}
    
     \caption{Robust Analysis over four systems on dialogue generation and text summarization.}
     \label{fig:rob_diag}
\end{figure*}

\subsection{Human Evaluation}

We recruit two human annotators from a professional data annotation team to evaluate our generated text in both informativeness and faithfulness.
The evaluations are conducted on text summarization and table-to-text generation.
All the annotators are first trained with the given criteria and demonstrations. 
After they are tested to be familiar with these criteria, we randomly select generated samples for annotation.
For each sample, the annotators are shown generated texts from  BART  and other three systems,  a  data augmentation method (SSTIA), a faithfulness  improvement method (CLIFF or Aug-plan), and our AdvSeq.
The annotators are asked to judge whether these generated texts are better than, tie with, or worse than the baseline model from the informativeness and faithfulness aspects, separately.
Evaluation results of 100 randomly selected samples from each dataset are reported in Table  \ref{tab:human_eval}.
Compared with other baselines, \textit{model trained with AdvSeq is more frequently rated as being more informative and more faithful.}

\section{Analysis}
\subsection{Robustness Analysis}
As demonstrated in Figure \ref{fig:rob_diag}, when testing on adversarial samples, models trained with AdvSeq degenerate the least and achieve the  best results in almost all the   metrics across tasks.
Especially for dialogue generation, Hit@1 performance does not decrease on adversarial samples for AdvSeq.
In summary, robustness analysis on three tasks validates that AdvSeq better improves robustness compared with other baselines.

\begin{table}[t]
    \centering
           \renewcommand{\arraystretch}{0.9}

     \setlength\tabcolsep{4.5pt}
    \begin{tabular}{l|ccccc}
    \Xhline{2\arrayrulewidth}
     & R-1 & R-2 & R-L &CC&QE\\
     \Xhline{2\arrayrulewidth}
   AdvGrad&\textbf{46.13}&\textbf{22.72}&\textbf{37.81}&23.49&\textbf{47.82}\\
   \textbf{w/o} $\delta_y$&45.81&22.39&37.47&\textbf{24.56}&47.56\\
    \textbf{w/o} $\delta_x$&46.08&22.65&37.70&23.85&47.75\\
    \textbf{w/o} $KL_S$&45.25&21.98&36.89&23.13&45.90\\
     BART&45.20&21.90&36.88&23.64&45.89\\
 \Xhline{2\arrayrulewidth}
    \end{tabular}
    \caption{Ablation study for AdvGrad on XSum.}
    \label{tab:vas_ab}
\end{table}

\subsection{AdvGrad Analysis}

\paragraph{Ablation Study} We study how each component affects the performance of AdvGrad  on XSum, reported in Table \ref{tab:vas_ab}.
We observe that, first, \textit{perturbations on  encoder and decoder both benefit AdvGrad.}
After removing  $\delta_x$  or $\delta_y$, every metric score drops (except CC). 
Second, \textit{KL is crucial for training with adversarial samples.}
Removing KL degrades faithfulness and informativeness into the similar performance with  BART (w/o KL vs BART).

\begin{figure}[t]
    \centering
    \includegraphics[scale=0.4276]{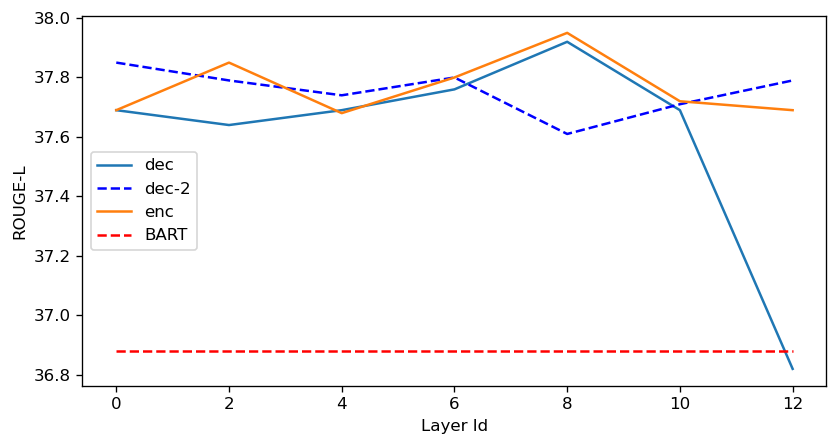}
    \caption{ROUGE-L scores of AdvGrad on XSum when training with perturbations on representations of different encoder (\textbf{enc}) or decoder (\textbf{dec})  layers. \textbf{dec-2} indicates training with two-step gradient ascent perturbations on the decoder .}
    \label{fig:attack_layer}
\end{figure}

\paragraph{Perturbed Layer}
We further study how perturbing representations from different layers  affect generation performance.
We conduct experiments on AdvGrad by gradually moving  $\delta_x$ or $\delta_y$ from word embeddings to high-layer representations of the encoder or decoder, separately.
In this procedure, the other side perturbation remains fixed on the word embedding.
The results are illustrated in Figure \ref{fig:attack_layer}, where the horizontal axis (Layer id) indicates the perturbed layer and the longitudinal axis reports the corresponding ROUGE-L score.\footnote{We test on even number layers.}
Because both BART encoder and decoder are composed of 12 transformer layers,  we denote layer  0  as the word embeddings and layer 12 as the output representations of the encoder or decoder.

From the results, we conclude that \textit{ in both encoder and decoder, perturbing the middle-layer representations achieves the best results.}
When layer id increases, ROUGE-L scores first increase to the peak at layer 8, and then decrease.
In particular,  when attacking on the output of the decoder (layer 12 of \textbf{dec}),  ROUGE-L score significantly drops because the perturbation only affects the final word prediction layer.

\paragraph{Multi-step Attack} We also study how multi-step gradient ascent (repeating  Eq.\ref{eq:adv_update}) affects generation.
We perform two-step gradient ascent on perturbing the decoder.
From Figure \ref{fig:attack_layer}, we can observe that comparing to one step ascent (\textbf{dec}), two-step ascent (\textbf{dec-2}) further benefits lower layers  (layer 0, 2, 4) but does not help or even worsen higher layers  (layer 6, 8, 10).
Overall, AdvGrad still has great potential.
Careful selection of permuted layers and gradient ascent steps will further boost its performance.

\subsection{AdvSwap Analysis}

We analyse how different settings in AdvSwap affect the overall  performance of AdvSeq.
In Table \ref{an:swap_all}, two factors are studied: selection of $\mathcal{Y}$, inference strategy for $\mathcal{X}$.
From the results, we can draw into  conclusion: \textit{More precise casualty inference lead to better performance, which depends on both inference strategies ($\mathcal{X}$) and the selection of $\mathcal{Y}$}.
In table-to-text, we can precisely infer $\mathcal{X}$ by word overlapping, since the correctly generated facts are mainly copied from the input.
While for  tasks like XSum, the summary is too abstract to infer precisely. 
Thus, changing the searching strategies (second rows in both Table \ref{ab:swap_ab}, \ref{an:swap_sum}) lead to drops in all metrics.
For  selecting $\mathcal{Y}$, using the whole $y$ with word overlapping in table-to-text will cause all the value tokens in table $x$  be selected into $\mathcal{X}$, which lead to worse performance.  
While in summarization, using global information in $y$ with gradient ranking  produces  the best performance.

\begin{table}[t]
    \centering
         \renewcommand{\arraystretch}{0.9}

     \begin{subtable}[h]{0.5\textwidth}
     \centering
      \setlength\tabcolsep{4.5pt}
    \begin{tabular}{ll|cc}
    \Xhline{2\arrayrulewidth}
     $\mathcal{Y}$&$\mathcal{X}$&BLEU&\textbf{PARENT}\\
     \Xhline{2\arrayrulewidth}
     
   Spans&Word Overlapping&30.37&57.33\\
Spans&Gradient Ranking&29.68&56.84\\
$y$ &Word Overlapping&
31.43&56.59\\
   
 \Xhline{2\arrayrulewidth}
    \end{tabular}
    \caption{Table-to-text, WIKIPERSON}
    \label{ab:swap_ab}
    
    \end{subtable}
          \renewcommand{\arraystretch}{0.9}

    \begin{subtable}[h]{0.5\textwidth}
    \centering
     \setlength\tabcolsep{3.8pt}
    \begin{tabular}{ll|cc}
     \Xhline{2\arrayrulewidth}
    $\mathcal{Y}$&$\mathcal{X}$&ROUGE-L&FactCC\\
    \Xhline{2\arrayrulewidth}
    $y$ &Gradient Ranking&37.81&24.67\\
    $y$&Word Overlapping&37.37&24.17\\
    Spans&Gradient Ranking&37.56&24.24\\

    \Xhline{2\arrayrulewidth}
    \end{tabular}
    
    \caption{Text summarization, XSum}
  \label{an:swap_sum}
    \end{subtable}
    \caption{Analysis on different settings of AdvSwap, where we report how the performance of AdvSeq changes when changing $\mathcal{X}$ and $\mathcal{Y}$.}
      \label{an:swap_all}
\end{table}

\section{Related Works}
\paragraph{Faithfulness in Text Generation}
Recently, a variety of works in text generation tasks have realized the severity of the unfaithful generation problem.
Improving the faithfulness of text generation has been an important and popular topic \cite{DBLP:journals/corr/abs-2203-05227}. 
Similar strategies have been proposed for different tasks.
For example, factual guidance is one of the  most popular methods.
\citet{DBLP:conf/aaai/CaoWLL18,li-etal-2020-leveraging-graph, wu-etal-2021-bass}  leverage guidance information like keywords and knowledge graph for faithful summarization; \citet{DBLP:conf/acl/RashkinRT020,DBLP:conf/aaai/WuGBZ0QKGHOD21}  respectively  guide response generation by evidence sentence and control phrases.
For data augmentation based methods, \citet{DBLP:journals/corr/abs-2112-01147} utilize entity swapping, while \citet{DBLP:conf/emnlp/Cao021} further construct various kinds of augmentation samples including entity regeneration, masking relation, etc.

\paragraph{Adversarial Learning}
Adversarial attack \cite{DBLP:journals/corr/GoodfellowSS14} searches for imperceptible perturbations to mislead neural networks.
Adversarial training \cite{DBLP:journals/corr/GoodfellowSS14}, derived from gradient-based adversarial attacks, trains the model with generated adversarial samples.
Besides extensive exploration in computer vision \cite{DBLP:conf/sp/Carlini017}, a large number of studies on adversarial learning for NLP emerge recently.
While most of them focus on  natural language understanding tasks \cite{DBLP:conf/acl/EbrahimiRLD18,DBLP:conf/emnlp/JiaL17,DBLP:conf/emnlp/LiMGXQ20}, some works also craft  adversarial samples for Seq2Seq models \cite{DBLP:conf/naacl/MichelLNP19,DBLP:conf/aaai/ChengYCZH20}.
For adversarial training, although several recent works apply gradient-based methods to fine-tune pre-trained NLU models \cite{DBLP:conf/iclr/ZhuCGSGL20,DBLP:conf/acl/JiangHCLGZ20}, adversarial analysis and learning for pre-trained Seq2Seq models is still lack of exploration.
\paragraph{Data Augmentation}
Data augmentation has been widely applied in almost every subarea in NLP.
Several studies propose universal data augmentation methods for text generation tasks.
Back-translation \cite{DBLP:conf/acl/SennrichHB16} constructs augmentation samples via auxiliary translation models,   which are widely used in improving the quality of machine translation.
R3F \cite{aghajanyan2020better} inject random noise on the word embeddings of the pre-trained language model for robust fine-tuning.
SSTIA \cite{xie2021target} applies a mix-up strategy on the target side of Seq2Seq to construct augmentation examples.
These methods mainly focus on improving the informativeness of generated text and lack of faithfulness analysis.

\section{Conclusion}
We are the first to conduct quantitative analysis on the robustness of pre-trained Seq2Seq models, and reveal the close connection between the robustness of Seq2Seq models and the informativeness and faithfulness for text generation tasks.
Through quantitative analysis, we point out that current SOTA pre-trained Seq2Seq model is still
not robust enough to tackle minor perturbations in the input.
To alleviate this problem, we propose AdvSeq to improve the faithfulness and informativeness of Seq2Seq models via enhancing their robustness.
Augmented by two types of adversarial samples during training, AdvSeq generates text with much better informativeness and faithfulness compared to strong baselines on text summarization, table-to-text, and dialogue generation.

\section*{Limitations}
\textbf{AdvGrad} \quad In  AdvGrad, we only  separately approximate the optimal perturbations for $x$ and $y$ with first order gradient.
 By exploring relations between  perturbations on encoder and decoder sides, more efficient and effective approximate solutions  of Eq.\ref{eq:adv_solve} may be given. \\
\textbf{Computational Cost} \quad  Compared with standard fine-tuning, AdvSeq constructs adversarial samples  with one extra backward pass and three extra forward passes.
AdvSeq requires approximately 4 times computation time and 1.7 times GPU memory usage for one batch than standard fine-tuning on BART.

\section*{Acknowledgments}
We thank the anonymous reviewers for their helpful comments on this paper. This work was partially supported by National Key Research and Development Project
(2019YFB1704002) and National Natural Science
Foundation of China (61876009).

\bibliography{anthology,custom}
\bibliographystyle{acl_natbib}
\newpage
\appendix

\section{Analysis Details}\label{app:robust_analyse}
\textbf{Evaluation Metrics} \quad For text summarization, we apply a Roberta classifier fine-tuned on MNLI dataset  as  $S_f(x,x')$ for predicting whether the modified sentences in $x'$ are factually consistent with their originals in $x$.

For table-to-text, because its inputs are tables of facts, it is hard  to modify them without changing their meaning, we only report how adversarial examples affect faithfulness by PARENT. 
PARENT-precision measures how much generated contents are covered by the input table or the reference text. 
PARENT-recall measures how much  contents in the input table or the reference text are mentioned in the generated context.\\
\textbf{Adversarial Sample} \quad Details of crafting adversarial samples are introduce in Algorithm \ref{app:craft_adv}.

\begin{algorithm}
\SetKwInOut{Input}{Require}\SetKwInOut{Output}{Output}
\Input{Original Sample $(x,y_{ref})$, Fine-tuned model $f_{\theta}$}
\SetKwFunction{FLoop}{Loop}
$y=f_\theta(x)$\\
$current\_bleu = BLEU (y,y_{ref})$\\
\If {$current\_bleu < 0.5$}{
    break
}
\Comment{Compute salient scores}\\
\For {$w_i$ in $x$}{ 
$s_i = f_\theta(y_{ref}|x)-f_\theta(y_{ref}|x\setminus w_i)$
}
$attack\_order = sorted(s)$\\
$x'\leftarrow x$\\
\Comment{Start attack}\\
\For {$i$ in $attack\_order$}{
    \If{$current\_bleu=0 $ or $Edit\_distance(x',x)>30$}{\textbf{return}  $x'$}
    \Comment{Search in top 10 nearest neighbors}\\
    \For {$\hat{w}_i$ in Nearest\_Neighbor$(w_i,10)$}{
    $x^{''}\leftarrow replace(x',\hat{w}_i)$\\
    $y'=f_{\theta}(x^{''})$\\
    $s' = BLEU(y',y_{ref})$\\
    \Comment{End condition}\\
    \If{$s' <current\_bleu$}{
        $current\_bleu \leftarrow s'$\\
        $x' \leftarrow x^{''}$\\
        $break$ \\}}
    }
\textbf{return} $x'$     
\caption{Adversarial Sample for Seq2Seq}
 \label{app:craft_adv}
\end{algorithm}

\section{AdvSeq}

Details of AdvSeq are illustrated in Algorithm \ref{app:advseq_al}.
Some cases of augmentation samples constructed by AdvSwap are demonstrated in Table \ref{tab:case_swap}.
From these samples we can see, AdvSwap mainly swaps subwords with their lexically or semantically similar candidates, which not only bring about diversity but also preserve the original meaning.

\begin{algorithm}
\SetKwInOut{Input}{Require}\SetKwInOut{Output}{Output}

\Input{Input sample $(x,y)$, model with parameters $\theta$}

\Comment{Objective function for the original sample}\\
$\delta_x, \delta_y \leftarrow U($-1e-2,+1e-2$)$\\
 $\mathcal{L}_o \leftarrow$ Eq.\ref{eq:KL_loss} with $(x,y,\delta_x, \delta_y)$\\
\Comment{Back-propagate $\mathcal{L}_o$ and save gradient of parameters $\theta$}\\
$g_0 \leftarrow \frac{1}{2}\nabla_\theta \mathcal{L}_o(x,y)$  \\
\Comment{AdvGrad}\\
 $\delta'_x \leftarrow \Pi_{||\delta_x||_F  \leq  \epsilon} (\delta_x + \alpha*g(\delta_x)/||g(\delta_x)||_F)$\\
 $ \delta'_y \leftarrow \Pi_{||\delta_y||_F  \leq  \epsilon} (\delta_y + \alpha*g(\delta_y)/||g(\delta_y)||_F)$\\
$\mathcal{L}_i \leftarrow$  Eq.\ref{eq:adv_grad} with $(x+\delta'_x, y+\delta'_y)$\\
 \Comment{
 Construct AdvSwap}\\
 Set $\mathcal{Y}, \mathcal{X}$ depend on the specific task\\
 Swap $\mathcal{X}$ in  $x$  on  Eq.\ref{eq:swap} to construct $x^{''}$\\
 $\mathcal{L}_e \leftarrow$ Eq.\ref{eq:adv_swap} with ($x^{''}, \mathcal{Y}$) \\
 \Comment{FreeLB training}\\
Back-propagate $\mathcal{L}_i + \mathcal{L}_e$   and accumulate gradient of parameters $\theta$:\\
 $g_1\leftarrow g_0+\frac{1}{2}\nabla_\theta(\mathcal{L}_i+\mathcal{L}_e)$\\
 \Comment{Update the $\theta$ with learning rate $\tau$}\\
 $\theta\leftarrow\tau g_1$
 \caption{AdvSeq}
 \label{app:advseq_al}
\end{algorithm}

\section{Datasets}
The statistic details of all the datasets for experiments all reported in Table \ref{tab:dataset}.
\begin{table}
    \centering
    \linespread{1.4}
    \begin{tabular}{lrrr}
      \Xhline{2\arrayrulewidth}
    Datasets&\#Train&\#Valid&\#Test  \\
    \Xhline{2\arrayrulewidth}
        XSum&204,045&11,332&11,334\\
        CNN/DM&287,227&13,368&11,490\\
       WIKIPERSON&250,186&30,487&29,982\\
        Dialogue NLI&10,013&968&12,376\\
     \Xhline{2\arrayrulewidth}
    \end{tabular}
    \caption{Statistics of datasets for evaluation.}
    \label{tab:dataset}
\end{table}

  \begin{table*}
\centering
    \begin{tabular}{p{0.15\textwidth}p{0.75\textwidth}}
    \specialrule{0em}{2pt}{2pt}
    \Xhline{2\arrayrulewidth}
    \multicolumn{2}{c}{\textbf{AdvSwap Cases  (XSum)}}\\
    \Xhline{2\arrayrulewidth}
   \textbf{Original $x$}: &The \textcolor{red}{accident} happened in 2009 during a play dress rehearsal at \textcolor{red}{the County Londonderry school }. The family had sued \textcolor{red}{the Western} Education and Library Board for alleged negligence. A judge ruled that the school had appropriately supervised the rehearsal . The judge told the court that he backed the family's account that the child had lost sight in his left eye because he was struck by \textcolor{red}{a girl} holding a wand , who was said to have been casting a spell \textcolor{red}{at  the time}. The boy , who can not \textcolor{red}{be} named for legal reasons , had been onstage with almost 200 other pupils getting ready for a performance . However , the judge held that teachers had properly assessed the wand and dismissed the claim . He said : " I do not consider the plaintiff has established any \textcolor{red}{fault on} the part of the defendant . "The judge praised the boy and his mother for the honesty of their evidence , but added there was " overwhelming evidence " that the girl who was holding the \textcolor{red}{fairy} wand was " timid and not likely to behave \textcolor{red}{in an inappropriate way} " .\\
     \textbf{AdvSwap $x^{''}$}&The \textcolor{red}{accidents} happened in 2009 during a play dress rehearsal at \textcolor{red}{The Londond school }. The family had sued \textcolor{red}{theWestern} Education and Library Board  for alleged negligence. A judge ruled that the school had appropriately supervised the rehearsal . The judge told the court that he backed the family's account that the child had lost sight in his left eye because he was strike \textcolor{red}{girls} holding a Wand , who was said to have been casting a spell \textcolor{red}{atthe time} . The boy , who can not \textcolor{red}{been} named for legal reasons , had been with almost 200 other pupils getting ready for a performance . However , the judge held that teachers had properly assessed the wand and dismissed the claim . He said : " I do not consider the plaintiff has established any \textcolor{red}{faultsOf} The defendant . "The judge praised the boy and hismother for the honesty of their evidence , but added there was " overwhelming evidence " that the girl who was holding the \textcolor{red}{Fairy} wand was " timid and not likely behaves \textcolor{red}{InAn Way} " .\\
    \Xhline{2\arrayrulewidth}
     \multicolumn{2}{c}{\textbf{AdvSwap Cases (WIKIPERSON)}}\\
     \Xhline{2\arrayrulewidth}
     \textbf{Original $x$}:&< Name ID > \textcolor{red}{Kevin Wheatley} < award received > Victoria Cross < date of birth > 13 March 1938 < date of death > 13 November 1965 < place of birth > \textcolor{red}{Surry Hills}, New South Wales < military branch > Australian Army\\
     \textbf{AdvSwap $x^{''}$}&  <NameID>\textcolor{red}{Kevin wheatleys}< award received > Victoria Cross < date of Birth> 13 March 1938 < date of death> 13 November 1965 < place of birth >\textcolor{red}{SurRY hills, }New South Welsh < military branch > Australian Army\\
     \Xhline{2\arrayrulewidth}
     \textbf{Original $x$}: &<Name ID >Albert Bierstadt < country of citizenship > Germany < country of citizenship > United States < date of birth > \textcolor{red}{January 7} 1830 < date of death > February 18 1902 < genre > \textcolor{red}{Landscape painting} < conflict > American Civil War \\
     \textbf{AdvSwap $x^{''}$}& <Name ID >Albert Bierstadt < country of citizenship > Germany < country of citizenship > United States< dates of Birth >\textcolor{red}{January7} 1830< dates of death>February18 1902 < genre > \textcolor{red}{Land Painting} < conflict> American Civil war\\
     \Xhline{2\arrayrulewidth}
      
     \end{tabular}
   
    \caption{Cases from AdvSwap}
     \label{tab:case_swap}
\end{table*}
     \begin{table*}
\centering
    \begin{tabular}{p{0.18\textwidth}p{0.75\textwidth}}
    \specialrule{0em}{3pt}{2pt}
    \Xhline{2\arrayrulewidth}
    \multicolumn{2}{c}{\textbf{Adversarial Attack Cases} (XSUM)}\\
    \Xhline{2\arrayrulewidth}
     \textbf{Original $x$}: &The \textcolor{blue}{Hangzhou} Internet Court opened on Friday and heard its first case - a copyright infringement dispute between an online writer and a web company.Legal agents in Hangzhou and Beijing accessed the court via their computers and the \textcolor{blue}{trial} lasted 20 minutes. ... In 2016, China began streaming some trials in more traditional courtrooms online in an apparent effort to boost the transparency of the legal system. ... .In some other countries, online portals to allow \textcolor{blue}{ people} to resolve legal disputes in cyber-space already exist.Canada's Civil Resolution Tribunal starting accepting claims for 5,000 (Â£3,000) or less in British Columbia in June\\
     \\
        \textbf{BART Output} &China's first internet court has opened its doors to the public for the first time in the world's largest city, state media report. .\\
        \textbf{AdvSeq Output} & 
        China has become the first country in the world to set up a court that allows people to sue each other via the internet , rather than in a traditional courtroom .\\
            \Xhline{2\arrayrulewidth}
        \textbf{Adversarial} $x'$:&The  Internet Court opened on Friday and heard its first case - a copyright infringement dispute between an online writer and a web company.Legal agents in Hangzhou and Beijing accessed the court via their computers and the \textcolor{blue}{experiment} lasted 20 minutes....."In 2016, China began streaming some trials in more traditional courtrooms online in an apparent effort to boost the transparency of the legal system....In some other countries, online portals to allow \textcolor{blue}{public} to resolve legal disputes in cyber-space already exist.Canada's Civil Resolution Tribunal starting accepting claims for 5,000 (Â£3,000) or less in British Columbia in June.\\
        \textbf{BART Output}&\textcolor{red}{A court in South Korea} has opened its doors to the public for the first time in the world's first internet habeas court, state media report..\\
        \textbf{AdvSeq Output}&China has become the first country in the world to open an internet - based court , which allows the public to take part in court cases via the internet.\\
                    \Xhline{2\arrayrulewidth}
 
  \end{tabular}
 \caption{A case from summarization}
   \label{case_1}
\end{table*}

    \begin{table*}
\centering
    \begin{tabular}{p{0.18\textwidth}p{0.75\textwidth}}
    \specialrule{0em}{3pt}{2pt}
    \Xhline{2\arrayrulewidth}
    \multicolumn{2}{c}{\textbf{Adversarial Attack Cases} (WIKIPERSON)}\\
    \Xhline{2\arrayrulewidth}
    \textbf{Original $x$}: &< Name ID > Adam Darski < date of birth > 10 \textcolor{blue}{June} 1977 < place of birth > Gdynia < country of citizenship > Poland < genre > Black metal < member of > \textcolor{blue}{Behemoth} (band)\\
    \\
    \textbf{BART Output}:& Adam Darski ( born 10 June 1977 in Gdynia Poland ) is a Polish musician best known as the lead vocalist of the Black metal band Behemoth (band) . Adam Darski was born on 10 June 1977 in Gdynia Poland . Behemoth (band) was formed in the early 2000s by Darski 's father Jan Darski and his brother Jan Darski .\\
    \textbf{AdvSeq Output:}&Adam Darski ( born 10 June 1977 in Gdynia Poland ) is a Polish musician best known as the lead vocalist of the Black metal band Behemoth (band) . Adam Darski was born in Gdynia Poland on 10 June 1977 . Behemoth (band) was formed in the early 1990s in Gdynia Poland.\\
    \Xhline{2\arrayrulewidth}
    \textbf{Adversarial} $x'$:&< Name ID > Damien Darski < date of birth > 10 \textcolor{blue}{April} 1977 < place of birth > Gdynia < country of citizenship > Poland < genre > Black metal < member of >\textcolor{blue}{ Dream} (band)\\
    \\
    \textbf{BART Output}:&Damien Darski ( born 10 April 1977 in Gdynia Poland ) is a Polish musician best known as the lead vocalist of the Black metal band Dream (band) . Damien Darski was born on 10 April 1977 in Gdynia Poland . \textcolor{red}{Dream (band) was formed in 1999 in Gdynia Poland. Dream (band) was formed in 1999 in Poland.}\\
     \textbf{AdvSeq Output:}&Damien Darski ( born 10 April 1977 in Gdynia Poland ) is a Polish musician best known as the lead vocalist of the Black metal band Dream (band) . Damien Darski was born on 10 April 1977 in Gdynia Poland . Dream (band) 's debut album Dream (band) was released in 2006 .\\
    \Xhline{2\arrayrulewidth}
    
\end{tabular}
 \caption{A case from table-to-text}

\label{case_2}

\end{table*}

\begin{table*}
\centering
    \begin{tabular}{p{0.17\textwidth}p{0.75\textwidth}}
    \specialrule{0em}{3pt}{2pt}
    \Xhline{2\arrayrulewidth}
    \multicolumn{2}{c}{\textbf{Adversarial Attack Cases} (WIKIPERSON)}\\
    \Xhline{2\arrayrulewidth}
    \textbf{Original $x$}: & < Name ID > Achille Raffray < occupation > \textcolor{blue}{Diplomat} < date of birth > 17 \textcolor{blue}{October} 1844 < date of death > 25 \textcolor{blue}{September} 1923 < country of citizenship > France < place of death > Rome < member of > Société entomologique de France\\
    \\
    \textbf{BART Output:}&Achille Raffray ( 17 October 1844 – 25 September 1923 ) was a France entomologist and Diplomat . Achille Raffray was born in Paris on 17 October 1844 . Achille Raffray died in Rome on 25 September 1923 . he was a member of the Société entomologique de France and the Société de France .\\
     \textbf{AdvSeq Output:}&Achille Raffray ( 17 October 1844 – 25 September 1923 ) was a France Diplomat and entomologist . he was a member of the Société entomologique de France. he was born in Paris and died in Rome .\\
        \Xhline{2\arrayrulewidth}
       \textbf{Adversarial} $x'$:&  < Name ID > Jacopo Raffray < occupation > \textcolor{blue}{Albright} < date of birth > 17 \textcolor{blue}{June} 1844 < date of death > 25 \textcolor{blue}{ Ninth }1923 < country of citizenship > France < place of death > Rome < member of > Société entomologique de France\\
       \\
       \textbf{BART Output:}&Jacopo Raffray ( 17 June 1844 – 25 Ninth 1923 Rome ) was a France entomologist . he was a member of the Société entomologique de France and the Société entomologique de France . raffray was born in Rio de Janeiro Brazil on 17 June 1844 and died in Rome on 25 Ninth 1923 . \textcolor{red}{he was a member of the Société entomologique de France and the Société astronomique de France . he studied at the École Normale Supérieure and the École pratique des hautes études en sciences sociales in Paris and became an Albright}\\
       \textbf{AdvSeq Output:}&Jacopo Raffray ( 17 Jane 1844 – 25 Ninth 1923 Rome ) was a France entomologist and Albright . he was a member of the Société entomologique de France. he was born in Rio de Janeiro Brazil and died in Rome Italy.\\
       \Xhline{2\arrayrulewidth}
\end{tabular}
 \caption{A case from table-to-text}

\label{case_3}

\end{table*}

\section{Case Study}
We demonstrate one case from XSum in Table \ref{case_1} and two other cases in Table \ref{case_2} and \ref{case_3}.

\end{document}